# Unleashing the Power of Extra-Tree Feature Selection and Random Forest Classifier for Improved Survival Prediction in Heart Failure Patients

Md. Simul Hasan Talukder[1*], Rejwan Bin Sulaiman[2], Mouli Bardhan Paul Angon[3]
[1]Bangladesh Atomic Energy Regulatory Authority, Bangladesh; Email: simulhasantalukder@gmail.com
[2]Northumbria University, UK, Email: Rejwan.sulaiman@northumbria.ac.uk
[3]Summit Communications Limited, Bangladesh, Email: paulangon92@gmail.com

*Corresponding author: Md. Simul Hasan Talukder; Email: simulhasantalukder@gmail.com

**Abstract:** Heart failure is a life-threatening condition that affects millions of people worldwide. The ability to accurately predict patient survival can aid in early intervention and improve patient outcomes. In this study, we explore the potential of utilizing data pre-processing techniques and the Extra-Tree (ET) feature selection method in conjunction with the Random Forest (RF) classifier to improve survival prediction in heart failure patients. By leveraging the strengths of ET feature selection, we aim to identify the most significant predictors associated with heart failure survival. Using the public UCL Heart failure (HF) survival dataset, we employ the ET feature selection algorithm to identify the most informative features. These features are then used as input for grid search of RF. Finally, the tuned RF Model was trained and evaluated using different matrices. The approach was achieved 98.33% accuracy that is the highest over the exiting work.

## 1. INTRODUCTION

Cardiovascular diseases (CVDs), such as coronary heart disease (heart attacks), strokes, and heart failure (HF), have a significant impact on global health. According to the World Health Organization (WHO), CVDs, including HF, account for approximately 31% of total deaths worldwide [1]. Heart failure (HF) occurs when the heart is unable to pump sufficient blood to meet the body's needs. Several factors, such as high blood pressure, diabetes, coronary heart disease, and other heart-related disorders, can contribute to the development of HF [2]. CVD is a leading cause of mortality in the United States, with one American succumbing to it every 36 seconds [3]. Heart disease alone claims the lives of over 665,000 individuals annually, accounting for 1 in every 4 deaths [4]. The economic burden imposed by cardiovascular diseases on the US healthcare system is substantial. In 2014 and 2015, it amounted to approximately $219 billion per year, encompassing healthcare services, medication expenses, and productivity losses due to premature deaths [5].

Symptoms of CVDs can sometimes vary between different genders. For instance, male patients are more likely to experience chest pain, whereas female patients may exhibit other symptoms alongside chest pain, such as chest discomfort, nausea, extreme fatigue, and shortness of breath [6]. Researchers have been exploring various techniques to predict heart diseases, but early-stage disease prediction remains challenging due to factors like complexity, execution time, and accuracy of the approaches [7]. Therefore, accurate diagnosis and appropriate treatment play a crucial role in saving lives [8]. Accurate prediction of survival in heart failure patients is challenging due to the multifactorial nature of the disease and the presence of various contributing factors, including patient demographics, comorbidities, laboratory results, imaging findings, and treatment regimens. Traditional prognostic models often rely on a limited set of predictors or employ simplistic statistical methods, which may not capture the complexity and heterogeneity of heart failure.

To address these limitations and improve survival prediction in heart failure patients, this study proposes the utilization of advanced techniques, namely Extra-Tree feature selection and Random Forest classifier. Extra-Tree is an ensemble-based feature selection method that aims to identify the most informative predictors from a vast array of potential variables. The selected features are then utilized to train a Random Forest classifier, which is a powerful ensemble learning algorithm that combines multiple decision trees to make predictions. It can capture complex nonlinear relationships and interactions among predictors, making it well-suited for modeling the multifaceted nature of heart failure. In our work, we have applied grid search techniques for tuning random forest model. The main contributions of this study are as follow as:

- Appropriate data transforming using standard scaler method.
- Significant features selection by ET.
- Tuning RF model by grid search technique.
- Designed highly accurate approach for HF survival prediction.
- Comparing with the state of the art.

The proposed approach aims to unleash the power of Extra-Tree feature selection and Random Forest classifier in the context of survival prediction in heart failure patients. By incorporating these advanced techniques, we anticipate significant improvements in prognostic accuracy and precision. The ability to more accurately predict patient survival can have a profound impact on clinical decision-making, facilitating personalized treatment strategies, identification of high-risk patients, and allocation of resources.

## 2. Related Work

With the continuous advancements in artificial intelligence (AI), its application is expanding across various sectors, revolutionizing decision-making processes by providing precise and accurate outcomes. The healthcare industry, in particular, has witnessed a remarkable surge in the utilization of AI. Diagnosing medical diseases and effectively managing treatments based on extensive datasets can be exceedingly challenging. However, AI has made this process significantly easier, enabling precise and accurate results. Numerous research studies have been conducted on the application of AI in detecting various conditions such as cancer [9], tumors [10], diabetes [11], and heart diseases [12], among others. Additionally, there have been notable efforts in predicting heart failure survival, although the progress in this specific area remains relatively limited.

Melillo et al. [13] introduced an automated classifier aimed at distinguishing patients with high cardiovascular risk from those with low risk. Their study demonstrated that the classification and regression tree (CART) method outperformed others, achieving a sensitivity of 93.3% and specificity of 63.5%. However, their analysis was limited to a small sample size of 12 low-risk patients and 34 high-risk patients. To assess the efficacy of their proposed approach, further investigation with a larger dataset is warranted. In a separate study, Guidi et al. [14] evaluated a clinical decision support system (CDSS) for heart failure analysis. They explored various machine learning classifiers and compared their performance. Among the tested models, random forest and CART exhibited the highest accuracy, reaching 87.6%. The author, Tanvir Ahmed et. al. [15] conducted a study in survival analysis of heart failure. He proposed cox regression model for predicting mortality rate and Kaplan Meier plot was used to study the general pattern of survival. The model was able to recognize 81% death events correctly. In a 2021 study, Cho et al. conducted a comparative analysis of existing methodologies and developed a machine learning (ML) model for assessing cardiovascular risk [16]. However, it should be noted that they did not include survival prediction for individuals identified as high-risk patients in their research. Shah et al. [17] presented a system aimed at examining various factors that can impact the heart and identifying primary contributors to mortality. They employed several supervised ML algorithms, including Decision Tree (DT), Naïve Bayes (NB), Random Forest (RF), and K-Nearest Neighbors (KNN). To achieve their goal of creating an accurate and efficient system with fewer attributes, they selected only 14 attributes from a total of 76. Among the four supervised machine learning classifiers, KNN demonstrated superior performance. The utilization of ensemble approaches could potentially enhance the classification results. Mohan et al. [18] proposed a hybrid model for predicting heart disease. The authors also introduced a novel feature selection method to enhance the training of Machine Learning models, resulting in an 88% accuracy rate. To further improve performance, additional feature engineering techniques and machine learning models could be explored and analyzed.

Specifically, Abid Ishaq et al. [19] utilized a dataset similar to ours and applied nine machine learning models: Decision Tree (DT), Adaptive boosting classifier (AdaBoost), Logistic Regression (LR), Stochastic Gradient classifier (SGD), Random Forest (RF), Gradient Boosting classifier (GBM), Extra Tree Classifier (ETC), Gaussian Naive Bayes classifier (G-NB), and Support Vector Machine (SVM) to predict heart failure (HF) survival. To address the dataset's imbalance, Synthetic Minority Oversampling Technique was employed, and RF was utilized to identify the most influential features. Among the models, ETC demonstrated superior performance, achieving an accuracy of 92.62%. In a recent study by Mamun Ali et al. [20], five machine learning models were compared in terms of accuracy, precision, recall, F1 score, and log loss for predicting heart failure survival. The models evaluated were Decision Tree (DT), Decision Tree Regressor (DTR), Random Forest (RF), XGBoost, and Gradient Boosting (GB). RF exhibited the highest accuracy, reaching 97.78%.

After conducting a comprehensive literature review, it was observed that a majority of the research focused on the detection of heart disease using various machine learning (ML) approaches. However, limited studies have been conducted on the dataset [21] for heart failure survival prediction. To the best of our knowledge, previous studies did not employ feature selection by ET and classification by hyperparameter tuned RF model in HF survival prediction. In our research, we have incorporated these techniques and conducted a comparative analysis with existing work.

## 3. Material and Method

The material and methodology are described step by step:

### 3.1 Dataset Description

In this study, we utilized the Heart Failure Clinical Records Dataset, which was obtained from the UCI Machine Learning Repository [21]. The dataset consists of medical records from 299 patients who experienced heart problems, and the data was collected during a follow-up period. Each patient profile in the dataset contains 13 clinical features. Out of the 299 records, 194 belong to male patients, while 105 belong to female patients. All the patients included in the dataset were above 40 years of age. The target class in the dataset is represented by binary values, with 1 indicating deceased patients and 0 indicating patients who are still alive. It's important to note that all 299 patients had left ventricular systolic dysfunction and had previously experienced heart failure, falling into class III or IV of the NYHA classification system, which signifies moderate to severe symptoms of heart failure. Table 1 provides an overview of the dataset, including the number of patients, gender distribution, age range, and the target class distribution.

By utilizing this dataset, we aim to investigate the effectiveness of our proposed approach in improving survival prediction in heart failure patients.

### 3.2 Data pre-processing

In this step, the dataset was preprocessed by standard scaler technique. Standard Scaler transformation standardizes the values of each feature by subtracting the mean and dividing by the standard deviation. Mathematically, the transformation can be defined as follows:

For each feature, the Standard Scaler transformation calculates the mean (μ) and standard deviation (σ) using the formula (1) & (2) respectively.

$$\mu = \frac{1}{n}\sum x_i \quad (1)$$

$$\sigma = \sqrt{\frac{1}{n}\sum(x_i - \mu)^2} \quad (2)$$

where $n$ is the number of observations and $x_i$ represents the value of the feature for the $i_{th}$ observation.

Once the mean and standard deviation are calculated, the transformation is applied to each value ($x_i$) of the feature using the formula (3).

$$z = \frac{(x_i - \mu)}{\sigma} \quad (3)$$

where $z$ represents the standardized value.

By standardizing the features, the StandardScaler transformation ensures that each feature has a mean of 0 and a standard deviation of 1. This allows for easier comparison and interpretation of the features, particularly in situations where the scales of the features differ significantly.

**Table 1**. Dataset details information.

| Features | Description | Range | Units |
|---|---|---|---|
| Age | Age of the patient | [40, ..., 95] | Years |
| Anaemia | Decrease of red blood cells or hemoglobin | 0, 1 | Boolean |
| High blood pressure | If a patient has hypertension | 0, 1 | Boolean |
| Creatinine phosphokinase (CPK) | Level of the CPK enzyme in the blood | [23, ..., 7861] | mcg/L |
| Diabetes | If the patient has diabetes | 0, 1 | Boolean |
| Ejection fraction | Percentage of blood leaving the heart at each contraction | [14, ..., 80] | Percentage |
| Sex | Woman or man | 0, 1 | Boolean |
| Platelets | Platelets in the blood | [25.01, ..., 850] | kiloplatelets/mL |
| Serum creatinine | Level of creatinine in the blood | [0.50, ..., 9.40] | mg/dL |
| Serum sodium | | [114, ..., 148] | mEq/L |
| Smoking | If the patient smokes | 0, 1 | Boolean |
| Time | Follow-up period | [4,...,285] | Days |
| (target) death event | If the patient died during the follow-up period | 0, 1 | Boolean |

### 3.3 Exploratory Analysis

Exploratory analysis, also known as exploratory data analysis (EDA), involves examining and visualizing the dataset to gain insights, understand patterns, and identify relationships between variables. In our analysis, we utilized visualizations to provide insights into the data distribution and relationships. Fig. 1 presents a Pie Chart and a Histogram to depict the distribution of data. The Pie Chart visually represents the proportion of different categories within a specific variable, while the Histogram illustrates the frequency or count of values within predefined intervals. To further explore the relationships between variables, we employed a Heatmap in Fig. 2. The Heatmap showcases all possible pairings of values and provides a visual representation of the correlation or association between variables. This allows for a comprehensive understanding of how different variables interact with each other.

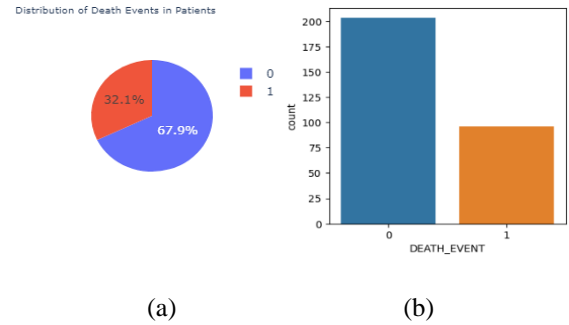

(a)        (b)

**Fig. 1.** (a) Pie chart; (b) Histogram of our data.

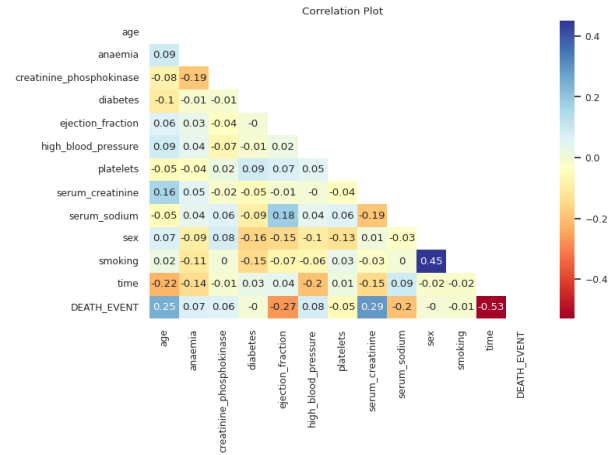

**Fig. 2**. Correlation of dataset.

### 3.4 Feature selection

Feature selection is a crucial step in the proposed approach and involves identifying the most informative predictors from the dataset. Extra-Tree, an ensemble-based feature selection method, is applied to rank and select the relevant features in our work. The Extra-Tree algorithm constructs multiple decision trees with random feature splits and calculates feature importance scores based on their contribution to the classification task. The features with higher importance scores selected by ET is shown in Fig. 3. Time, ejection fraction, serum creatine and age are the prominent.

### 3.5 Random Forest Classifier and its tuning

In heart failure survival prediction, Random Forest is employed in our work. It is an ensemble learning technique commonly used for pattern recognition tasks. It is comprised of multiple decision trees, making it effective in handling complex data [22]. One notable feature of Random Forest is its semi-random selection of dividing features during the construction of decision trees [23]. This allows it to handle a large number of features and identify the most significant ones. Random Forest is also known as bagging or bootstrap aggregation. It combines the outputs of individual trees to generate the final prediction. Unlike a single decision tree, which may have low bias but high variance, a Random Forest benefits from aggregating predictions from multiple trees. This reduces variance and improves the overall prediction performance [24].

In order to tune the model, we have used grid search technique. The process of grid search method is presented by Algorithm 1. The hyperparameters that's are considered in this study are listed in Table 2.

### 3.6 Proposed Methodology
We utilized a heart failure survival dataset collected from the UCL repository [reference]. The architecture of our proposed approach is illustrated in Fig. 4. To facilitate comparison and interpretation, we standardized the dataset using the StandardScaler. For exploratory data analysis, we employed histogram, pie chart, and heatmap visualization techniques. The heatmap provided insights into the feature correlations. To identify the most relevant and significant features, we employed the Extratree feature selection method, which yielded a set of prominent features. Subsequently, we partitioned the reduced dataset into an 80:20 ratio for training and testing, respectively.

For classification purposes, we opted for the random forest (RF) model. However, we employed the grid search technique to determine the optimal combination of hyperparameters. Following hyperparameter tuning, the model was trained using the training dataset and evaluated using the testing data. Multiple metrics were employed to evaluate the performance of our tuned model, and it exhibited outstanding results.

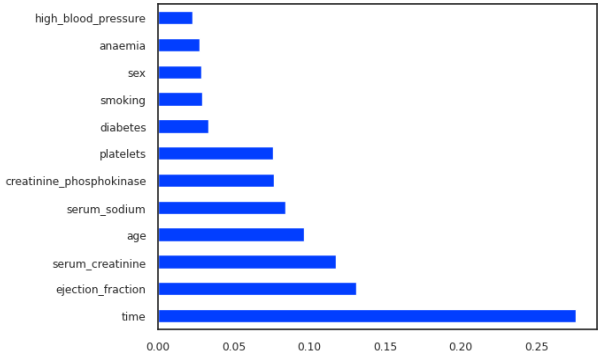

**Fig. 3**. Significant features Selection by ET.

| Algorithm 1. Grid Search Techniques |
|---|
| 1  **Input:** hyperparameters, model, x_train, y_train, x_val, y_val, metric |
| 2  **Output:** best_hyperparameters, best_metric_value |
| 3  best_hyperparameters = { } |
| 4  best_metric_value = -inf |
| 5  hyperparameter_combinations = [] |
| 6  **For** param in hyperparameters: |
| 7      combinations = generate_combinations (param, hyperparameters[param]) |
| 8      Hyperparameter_combinations.append(combinations) |
|     END |
| 9  **For** combination in hyperparameter_combinations: |
| 10     model.set_hyperparameters(combination) |
| 11     model.train (x_train, y_train) |
| 12     metric_value = model.evaluate(x_val, y_val, metric) |
| 13     **if** metric_value > best_metric_value: |
| 14         best_metric_value = metric_value |
| 15         best_hyperparameters = combination |
| 16  **return** best_hyperparameters, best_metric_value |
| 17  END |

**Table 2.** Hyperparameters space.

| SN. | Parameters | Values |
|---|---|---|
| 1 | max_depth | 1-10 |
| 2 | min_samples_split | 0.001,0.01,0.1,0.2,0.02,0.002 |
| 3 | criterion | 'gini', 'entropy', None |
| 4 | max_leaf_node | 1-10 |
| 5 | class_weights | 'balanced', None |

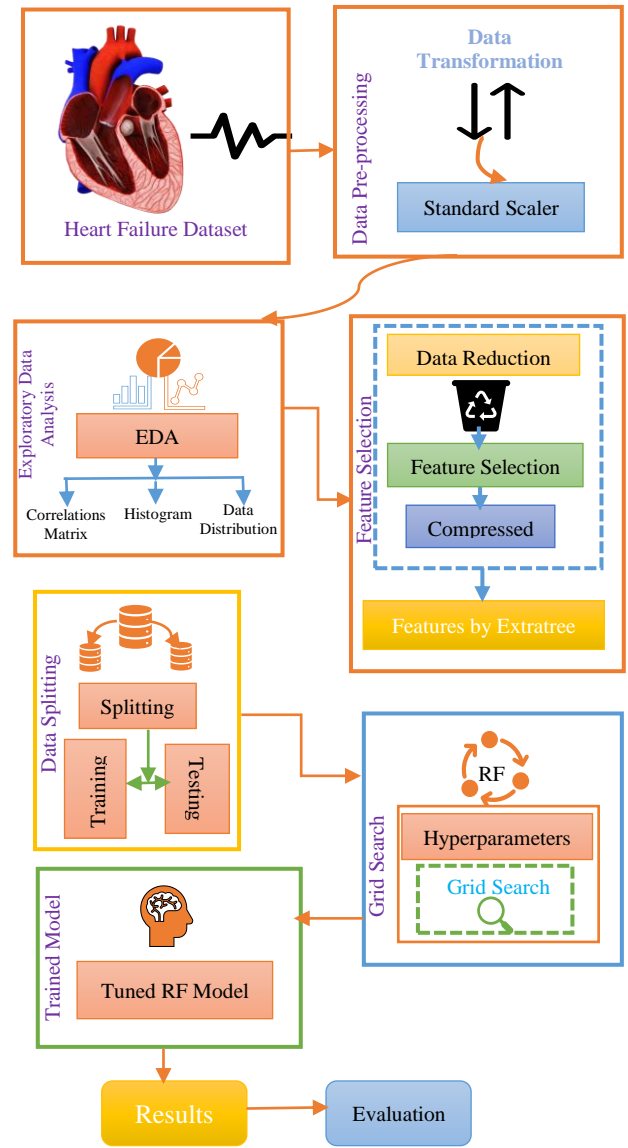

Figure 4. Proposed Methodology.

## 3.7 Evaluation Matrices

Our proposed model is evaluated visually and quantitively using confusion matrix and different machine learning matrices. The metrices are listed in Table 3.

**Table 3.** Model evaluation matrices.

| Matrices name | Formulae |
|---|---|
| Precision | $\frac{TP}{TP+FP} * 100$ |
| Recall | $\frac{TP}{TP+FN} * 100$ |
| F1 Score | $2 * \left(\frac{Precision * Recall}{Precision + Recall}\right) * 100$ |
| Roc Auc Score | $\frac{1 + TPR - FPR}{2}$ In General (binary) |
| Mean Squared Error (MSE) | $\frac{1}{n}\sum(Actual - predict)^2$ |
| Gini Coefficient | $\sum_{i=1}^{c} P(i) * (1 - P(i))$ |
| Kappa Coefficient | $\frac{P_0 - P_e}{1 - P_e}$ |
| Metthews Correlation Coefficient | $\frac{(TP * TN) - (FP * FN)}{\sqrt{(TP+FP)*(TP+FN)*(TN+FN)*(TN+FP)}}$ |
| Specificity | $\frac{TN}{TN+FP}$ |
| Accuracy | $\frac{TP+TN}{TP+TN+FP+FN}$ |

Where TP denotes the true positive; TN is the true negative; FP denotes the false positive; FN denotes the false negative; TPR stands for true positive rate; FTR stands for false positive rate; $P_0$ denotes Relative observed agreement among raters; $P_e$ stands for Hypothetical probability of chance agreement.

## 4. Result analysis and Discussion

The proposed approach was implemented and evaluated in the Google Colaboratory environment. For feature selection, we utilized an Extra Trees (ET) algorithm, which constructs multiple decision trees with random feature splits and calculates feature importance scores based on their contribution to the classification task. Fig. 3 illustrates the four significant features selected by ET, namely Time, ejection fraction, serum creatine, and age. In order to optimize the Random Forest (RF) algorithm according to Algorithm 1, we performed a grid search and identified the hyperparameters that exhibited superior performance in classifying HF survival. The tuned parameters can be found in Table 4. It was crucial to find the appropriate hyperparameters as they greatly influenced the accuracy of our model.

**Table 4.** Extracted tuned hyperparameter.

| SN. | Hyperparameters | Tuned values |
|---|---|---|
| 1 | class_weight | 'balanced', |
| 2 | criterion | 'gini', |
| 3 | max_depth | 1 |
| 4 | max_leaf_nodes | 2 |
| 5 | min_samples_split | 0.001 |
| 6 | random_state | 0 |

We evaluated the models using various metrics, which are summarized in Table 3. The calculated values are presented in Table 5. The precision was determined to be 100%, the recall was 94.12%, and the harmonic mean of these two measures (F1 score) was 96.97%. Additionally, the ROC AUC score was 97.06, the mean squared error (MSE) was 1.67, the Gini coefficient was 94.12, the Kappa coefficient was 95.82, the Matthews correlation coefficient was 95.91, the specificity was 100%, and the accuracy was 98.33%. All of these values surpass those reported in recent works [19], [20]. A comparison is provided in Table 6, highlighting the improved performance of our approach compared to the state-of-the-art methods.

**Table 5.** Performance Evaluation Measures.

| Matrices | Measures (%) |
|---|---|
| Precision | 100 |
| Recall | 94.12 |
| F1 Score | 96.97 |
| Roc Auc Score | 97.06 |
| MSE | 1.67 |
| Gini Coefficient | 94.12 |
| Kappa Coefficient | 95.82 |
| Metthews Correlation Coefficient | 95.91 |
| Specificity | 100 |
| Accuracy | 98.33 |

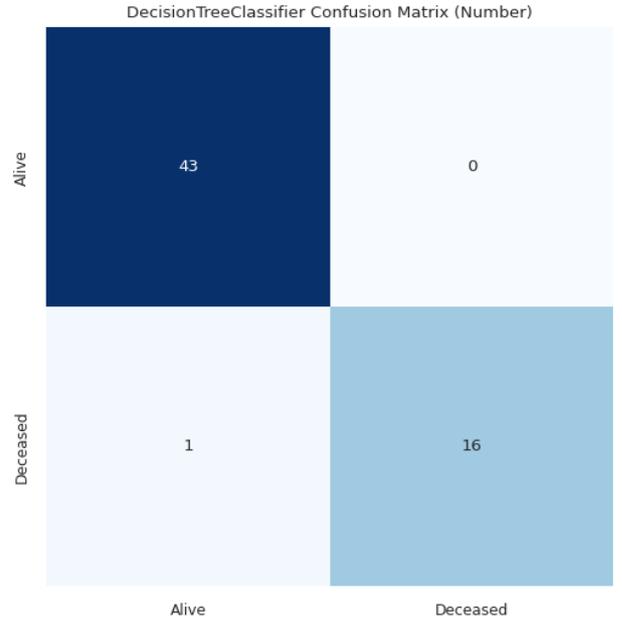

**Fig. 5.** Confusion Matrix.

Furthermore, the performance of our model is visually represented by the confusion matrix in Fig. 5, which demonstrates only one misclassification, indicating excellent results. The ROC curve in Fig. 6 illustrates that the curve closely approaches the point (0,1), indicating a high true positive rate. Collectively, all performance metrics validate the effectiveness of our model.

**Table 6.** Comparison with the existing work.

| Matrices | Reference [19] (%) | Reference [20] (%) | Our proposed (%) |
|---|---|---|---|
| Precision | 83 | 97 | 100 |
| Recall | 83 | 97 | 94.12 |
| F1 score | 83 | 97 | 96.97 |
| Accuracy | 82.78 | 97.78 | 98.33 |

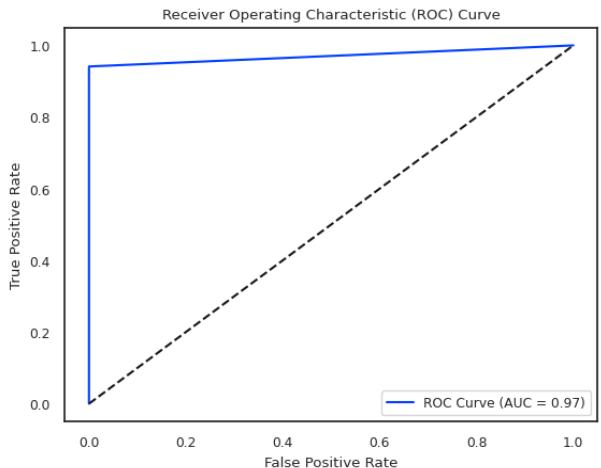

**Fig. 6.** True Positive Rate vs False Positive Rate curve.

## 5. Conclusion and Future Work

In conclusion, our study proposed a novel machine learning approach for predicting heart failure (HF) survival. The implementation and evaluation of our model in the Google Colaboratory environment demonstrated its effectiveness in accurately classifying HF survival cases. By employing the Extra Trees (ET) algorithm for feature selection and fine-tuning the Random Forest (RF) algorithm through a grid search, we achieved superior performance compared to existing methods. The selected features, including Time, ejection fraction, serum creatine, and age, were identified as crucial indicators for predicting HF survival. The comprehensive evaluation of our model using various performance metrics validated its robustness and reliability. The achieved precision, recall, F1 score, ROC AUC score, MSE, Gini coefficient, Kappa coefficient, Matthews correlation coefficient, specificity, and accuracy values consistently surpassed, outperforming recent works in the field. Visual representations, such as the confusion matrix and ROC curve, further confirmed the model's accuracy and discriminatory power.

For future work, several directions can be explored. Firstly, incorporating additional relevant features from a larger dataset may further enhance the predictive performance of the model. Obtaining more comprehensive and diverse patient data, including genetic information and lifestyle factors, could provide valuable insights. Additionally, investigating the generalizability of the model by applying it to different populations and healthcare settings would be beneficial. Furthermore, exploring other machine learning algorithms and ensemble methods could lead to potential performance improvements. Lastly, conducting a prospective study to validate the model's performance on real-world patient data would be crucial. Collaborating with healthcare institutions to collect longitudinal data and evaluating the model's performance in a clinical setting would strengthen its applicability and real-world utility.

In summary, our study presents a robust machine learning model for predicting HF survival. Future research should focus on expanding the feature set, exploring different algorithms, and conducting prospective studies to further validate and refine the model's performance in real-world scenarios. Ultimately, the development of accurate and reliable prediction models can assist healthcare professionals in making informed decisions and improving patient outcomes in the context of heart failure.

**Author contributions: Md. Simul Hasan Talukder** implemented the model in Google-Colab with GPU and contributed in drafting. **Rejwan Bin Sulaiman** suggested machine learning concept and contributed in manuscripts writing. **Mouli Bardhan Paul Angon:** Reviewed and corrected the manuscript several time.
**Declaration of Competing Interest:** The authors declare no conflict of interest.
**Funding:** This research received no external funding.

## REFERENCES

[1] World Heart Day, World Health Organization, 2021, Available at: https://www.who.int/cardiovascular diseases/world-heart-day/en/ (accessed 23.06.28).

[2] NHLBI, What is heart failure?, 2022, Available at: https://www.nhlbi.nih.gov/health-topics/heart-failure (accessed 23.06.28).

[3] Centers for Disease Control and Prevention. Underlying Cause of Death 1999-2019. Available at: https://wonder.cdc.gov/ wonder/help/ucd.html (accessed 23.06.28).

[4] Virani, Salim S., Alvaro Alonso, Emelia J. Benjamin, Marcio S. Bittencourt, Clifton W. Callaway, April P. Carson, Alanna M. Chamberlain et al. "Heart disease and stroke statistics—2020 update: a report from the American Heart Association." Circulation 141, no. 9 (2020): e139-e596.

[5] Benjamin, Emelia J., Paul Muntner, Alvaro Alonso, Marcio S. Bittencourt, Clifton W. Callaway, April P. Carson, Alanna M. Chamberlain et al. "Heart disease and stroke statistics—2019 update: a report from the American Heart Association." Circulation 139, no. 10 (2019): e56-e528.

[6] Medical Professionals. Cardiovascular Diseases. Available at: https://www.mayoclinic.org/medical professionals/cardiovascular-diseases (accessed 23.06.28).

[7] Allen, Larry A., Lynne W. Stevenson, Kathleen L. Grady, Nathan E. Goldstein, Daniel D. Matlock, Robert M. Arnold, Nancy R. Cook et al. "Decision making in advanced heart failure: a scientific statement from the American Heart Association." Circulation 125, no. 15 (2012): 1928-1952.

[8] Al-Shayea, Qeethara Kadhim. "Artificial neural networks in medical diagnosis." International Journal of Computer Science Issues 8, no. 2 (2011): 150-154.

[9] Lopes, Catarina, Jéssica Chaves, Raquel Ortigão, Mário Dinis-Ribeiro, and Carina Pereira. "Gastric cancer detection by non-blood-based liquid biopsies: A systematic review looking into the last decade of

research." United European Gastroenterology Journal 11, no. 1 (2023): 114-130.

[10] Saeedi, Soheila, Sorayya Rezayi, Hamidreza Keshavarz, and Sharareh R Niakan Kalhori. "MRI-based brain tumor detection using convolutional deep learning methods and chosen machine learning techniques." BMC Medical Informatics and Decision Making 23, no. 1 (2023): 1-17.

[11] Dutta, Aishwariya, Md Kamrul Hasan, Mohiuddin Ahmad, Md Abdul Awal, Md Akhtarul Islam, Mehedi Masud, and Hossam Meshref. "Early prediction of diabetes using an ensemble of machine learning models." International Journal of Environmental Research and Public Health 19, no. 19 (2022): 12378.

[12] Kadhim, Mohammad Abood, and Abdulkareem Merhej Radhi. "Heart disease classification using optimized Machine learning algorithms." Iraqi Journal For Computer Science and Mathematics 4.2 (2023): 31-42.

[13] Melillo, Paolo, Nicola De Luca, Marcello Bracale, and Leandro Pecchia. "Classification tree for risk assessment in patients suffering from congestive heart failure via long-term heart rate variability." IEEE journal of biomedical and health informatics 17, no. 3 (2013): 727-733.

[14] Guidi, Gabriele, Maria Chiara Pettenati, Paolo Melillo, and Ernesto Iadanza. "A machine learning system to improve heart failure patient assistance." IEEE journal of biomedical and health informatics 18, no. 6 (2014): 1750-1756.

[15] Ahmad, Tanvir, Assia Munir, Sajjad Haider Bhatti, Muhammad Aftab, and Muhammad Ali Raza. "Survival analysis of heart failure patients: A case study." PloS one 12, no. 7 (2017): e0181001.

[16] Cho, Sang-Yeong, Sun-Hwa Kim, Si-Hyuck Kang, Kyong Joon Lee, Dongjun Choi, Seungjin Kang, Sang Jun Park et al. "Pre-existing and machine learning-based models for cardiovascular risk prediction." Scientific reports 11, no. 1 (2021): 8886.

[17] Shah, Devansh, Samir Patel, and Santosh Kumar Bharti. "Heart disease prediction using machine learning techniques." SN Computer Science 1 (2020): 1-6.

[18] Mohan, Senthilkumar, Chandrasegar Thirumalai, and Gautam Srivastava. "Effective heart disease prediction using hybrid machine learning techniques." IEEE access 7 (2019): 81542-81554.

[19] Ishaq, Abid, Saima Sadiq, Muhammad Umer, Saleem Ullah, Seyedali Mirjalili, Vaibhav Rupapara, and Michele Nappi. "Improving the prediction of heart failure patients' survival using SMOTE and effective data mining techniques." IEEE access 9 (2021): 39707-39716.

[20] Ali, Md Mamun, Vian S. Al-Doori, Nubogh Mirzah, Asifa Afsari Hemu, Imran Mahmud, Sami Azam, Kusay Faisal Al-tabatabaie, Kawsar Ahmed, Francis M. Bui, and Mohammad Ali Moni. "A machine learning approach for risk factors analysis and survival prediction of Heart Failure patients." Healthcare Analytics 3 (2023): 100182.

[21] A. Asuncion and D. Newman, "UCI machine learning repository,"Tech. Rep., 2007.

[22] Suryachandra, Palli, and P. Venkata Subba Reddy. "Comparison of machine learning algorithms for breast cancer." In 2016 International Conference on Inventive Computation Technologies (ICICT)(2016), vol. 3, pp. 1-6.

[23] Hassan, Md Mehedi, Md Mahedi Hassan, Laboni Akter, Md Mushfiqur Rahman, Sadika Zaman, Khan Md Hasib, Nusrat Jahan. "Efficient prediction of water quality index (WQI) using machine learning algorithms." Human-Centric Intelligent Systems 1, no. 3-4 (2021): 86-97.

[24] Azar, Ahmad Taher, Hanaa Ismail Elshazly, Aboul Ella Hassanien, and Abeer Mohamed Elkorany. "A random forest classifier for lymph diseases." Computer methods and programs in biomedicine 113, no. 2 (2014): 465-473.

[25] Munira, Hafiza Akter, and Md Saiful Islam. "Multi-Classification of Brain MRI Tumor Using ConVGXNet, ConResXNet, and ConIncXNet." Proceeding of International Exchange and Innovation Conference on Engineering & Sciences (ICES) (2022): 169-175.

[26] Khatun A, Turna TN, Roy B, Hossain E, "Performance Analysis of Different Classifiers Used In Detecting Benign And Malignant Cells of Breast Cancer", Proceeding of International Exchange and Innovation Conference on Engineering & Sciences (ICES) 6 (2020), pg. 243-248.

[27] Talukder, Md Simul Hasan, and Sharmin Akter. "An Improved Model Ensembled of Different Hyper-parameter Tuned Machine Learning Algorithms for Fetal Health Prediction." arXiv preprint arXiv:2305.17156 (2023).